\documentclass[letterpaper, 10pt, journal, twoside]{IEEEtran}
\usepackage{cite}
\usepackage{bbm}
\usepackage{multirow}
\usepackage{graphicx} 
\usepackage{booktabs}
\usepackage[namelimits]{amsmath} 
\usepackage{amssymb}             
\usepackage{amsfonts}    
\usepackage{dsfont}
\usepackage{mathrsfs}           
\usepackage{enumitem}
\usepackage{lipsum}
\usepackage{url}
\usepackage{xcolor}
\usepackage{algorithm}
\usepackage{algpseudocode}
\usepackage{hyperref}
\usepackage[capitalize]{cleveref}
\hypersetup{
    colorlinks=true,
    linkcolor=blue,
    filecolor=magenta,      
    urlcolor=cyan
}

\begin{document}

%
\title{Learning Online Belief Prediction for Efficient POMDP Planning in Autonomous Driving}
%
%
%

\author{Zhiyu Huang$^{1}$,
        Chen Tang$^{2}$,
        Chen Lv$^{1}$,
        Masayoshi Tomizuka$^{3}$,
        and Wei Zhan$^{3}$
\thanks{Manuscript received: December 19, 2023; Revised May 4, 2024; Accepted June 9, 2024.}
\thanks{This paper was recommended for publication by Editor Hanna Kurniawati upon evaluation of the Associate Editor and Reviewers' comments.} 
\thanks{$^{1}$Z. Huang and C. Lv are with the School of Mechanical and Aerospace Engineering, Nanyang Technological University, Singapore. This work was done during Z. Huang's visit to the University of California, Berkeley. {\tt zhiyu001@e.ntu.edu.sg, lyuchen@ntu.edu.sg}}
\thanks{$^{2}$C. Tang is with the Department of Computer Science at the University of Texas, Austin, TX, USA. {\tt chen.tang@austin.utexas.edu}}
\thanks{$^{3}$M. Tomizuka and W. Zhan are with the Department of Mechanical Engineering at the University of California, Berkeley, CA, USA. {\tt \{tomizuka, wzhan\}@berkeley.edu}}
\thanks{Digital Object Identifier (DOI): see top of this page.}
}

%
%

\markboth{IEEE Robotics and Automation Letters. Preprint Version. Accepted June, 2024}
{HUANG \MakeLowercase{\textit{et al.}}: Learning Online Belief Prediction for Efficient POMDP Planning in Autonomous Driving} 

%


\maketitle

\begin{abstract}
Effective decision-making in autonomous driving relies on accurate inference of other traffic agents' future behaviors. To achieve this, we propose an online belief-update-based behavior prediction model and an efficient planner for Partially Observable Markov Decision Processes (POMDPs). We develop a Transformer-based prediction model, enhanced with a recurrent neural memory model, to dynamically update latent belief state and infer the intentions of other agents. The model can also integrate the ego vehicle's intentions to reflect closed-loop interactions among agents, and it learns from both offline data and online interactions. For planning, we employ a Monte-Carlo Tree Search (MCTS) planner with macro actions, which reduces computational complexity by searching over temporally extended action steps. Inside the MCTS planner, we use predicted long-term multi-modal trajectories to approximate future updates, which eliminates iterative belief updating and improves the running efficiency. Our approach also incorporates deep Q-learning (DQN) as a search prior, which significantly improves the performance of the MCTS planner. Experimental results from simulated environments validate the effectiveness of our proposed method. The online belief update model can significantly enhance the accuracy and temporal consistency of predictions, leading to improved decision-making performance. Employing DQN as a search prior in the MCTS planner considerably boosts its performance and outperforms an imitation learning-based prior. Additionally, we show that the MCTS planning with macro actions substantially outperforms the vanilla method in terms of performance and efficiency.
\end{abstract}


%
\IEEEpeerreviewmaketitle

\section{Introduction}
\IEEEPARstart{D}{ecision}-making under uncertainties is crucial for the safety of autonomous driving systems. In particular, human traffic participants' behaviors are a primary source of uncertainties, which imposes significant challenges to the safe navigation of autonomous vehicles (AVs) in real-world scenarios. The Partially Observable Markov Decision Process (POMDP) \cite{brown2020taxonomy} offers a mathematically sound framework to address this problem. However, most POMDP planners in autonomous driving have limited capabilities, as they only represent the hidden states of other agents with specific semantic meanings \cite{sunberg2022improving, hubmann2018automated, fischer2022guiding}, which restricts their capability and scalability in complex real-world situations. To overcome this issue, several studies propose enhancing POMDP planning with learned neural network policy and value priors \cite{moss2023betazero, cai2019lets, cai2022closing}. However, within the context of autonomous driving, learning the state transition function (i.e., predicting the future actions of other agents) is the most challenging part of POMDP planning. Therefore, we aim to integrate POMDP planning with deep learning-based prediction models and develop an online behavior prediction model for effective planning.

\begin{figure}
    \centering
    \includegraphics[width=0.95\linewidth]{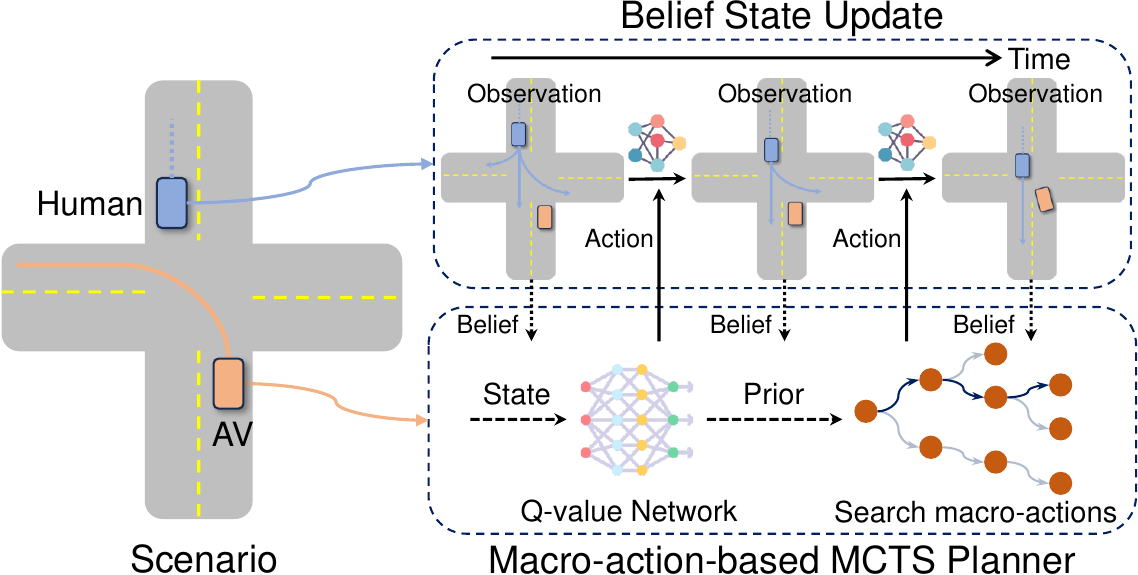}
    \caption{Illustration of our proposed planning approach. We utilize a neural memory-based belief update model to continually update other human agents' intentions over time based on new observations and the AV's actions. A macro-action-based MCTS planner, guided by a learned Q-value function, searches for the approximately optimal action based on the current belief state.}
    \label{fig:1}
    \vspace{-0.5cm}
\end{figure}

One challenge is adapting the prediction model to an online and closed-loop setting. We propose a neural memory-based belief update model that optimizes closed-loop prediction performance and learns through online interactions. Specifically, our proposed model uses a Transformer-based encoder to map the observation to latent space. At each time step in online training or testing, we utilize a gated recurrent unit (GRU) model to update the latent belief state of the AV about each tracked agent by considering their last latent states, the current latent observations, and the intention of the ego agent. We then employ a decoder to map the latent belief state back into the explicit belief state, represented by distributions of intentions (long-term trajectories) for other agents. Our neural memory model can significantly improve the temporal consistency and probability estimation accuracy of behavior prediction models in online testing. Our model is designed to capture the dynamic interactions between the AV and other agents, as opposed to open-loop conditional prediction models \cite{ tolstaya2021identifying, huang2023conditional}.

Another challenge lies in developing a computationally efficient POMDP planner for AVs. We adopt the Monte-Carlo tree search (MCTS) algorithm \cite{chekroun2023mbappe, li2022efficient} and incorporate several enhancements. First, we leverage a macro-action-based method that searches over action sequences or motion primitives \cite{zhou2023towards, de2016monte}, allowing a more in-depth search within a limited computation budget. Additionally, we use predicted multi-modal long-term trajectories from the online prediction model to approximate future transitions for other agents within the planning horizon, which improves computational efficiency while ensuring planning performance. Lastly, we train a deep Q-value network based on the environment reward feedback during the online learning process, which is used as the search prior for the MCTS planner.

In summary, we focus on the decision-making under uncertainty problem for AVs, and we have proposed several learning-based enhancements for effective POMDP planning. The core idea of our approach is illustrated in \cref{fig:1}, and the main contributions are outlined below:
\begin{enumerate}[label=C\arabic*)]
\item We propose a neural memory-based belief update model that provides online and closed-loop agent behavior prediction for AV planning. Refer to \cref{online model}.

\item We introduce a macro-action-based MCTS planning method for AVs, which integrates the online prediction model for future approximation and a Q-value function network that acts as a heuristic guide. See \cref{mcts}.

\item We establish an online learning framework of belief update model and Q-value network and validate our proposed method with a real-world driving dataset and simulated driving environment. Refer to \cref{learning}.
\end{enumerate}

\section{Related Work}
\textbf{POMDP and MCTS in Autonomous Driving}.
POMDP is a mathematically principled framework for decision-making under uncertainty, which is increasingly being used in AV planning. However, conventional methods are limited to handling simple POMDPs with semantic unobservable properties, such as parameters of driving behavior models \cite{sunberg2022improving}, possible routes to follow \cite{hubmann2018automated}, cooperation levels \cite{fischer2022guiding}, or intelligence levels \cite{dai2023game}. MCTS algorithms have been identified as an effective approach to solving POMDPs \cite{sunberg2018online}, and we convert POMDP to belief-state MDP \cite{moss2023betazero} to solve it more efficiently. Using learned models, including the transition dynamics, policy, and value functions, can significantly improve the performance and scalability of MCTS \cite{schrittwieser2020mastering}. \cite{cai2019lets} and \cite{cai2022closing} proposed using learned policy and value functions to guide the belief tree search for solving crowd navigation tasks. \cite{fischer2022guiding} proposed using neural networks to estimate the values of leaf nodes and guide the tree search for interactive merging scenarios. However, existing models still presume semantic belief parameters in POMDPs and fail to generalize to complicated situations. Our work addresses this gap by modeling the belief state as general behaviors or intentions of other road users and uses learned models to approximate future transitions, which is critical yet challenging in autonomous driving tasks. Although previous studies have incorporated learning-based behavior prediction models into MCTS planning \cite{chekroun2023mbappe, li2022efficient}, they typically rely on MDP settings without online belief updates, and the prediction models are often trained offline. Our approach develops an online prediction model that utilizes a neural network for online belief updating (\textcolor{red}{\textbf{C1}}). Additionally, we leverage a learned deep Q-network as a search heuristic in the MCTS planner, which shows significant improvements in planning performance (\textcolor{red}{\textbf{C3}}).


Employing recurrent networks to track and update (latent) belief states \cite{igl2018deep, jonschkowski2018differentiable} is an effective approach for POMDPs. In our AV planning approach, we utilize a recurrent GRU network to continually update the latent belief state concerning the behaviors of other traffic participants, by distributing the ego agent's belief state across these gents. Furthermore, to deal with a high-dimensional and continuous search space, we implement a macro-action-based method (\textcolor{red}{\textbf{C2}}) that searches over action sequences instead of raw control actions \cite{zhou2023towards}, which effectively reduces search depth. Although previous studies have proposed learning-based models in POMDPs and neural belief updaters, adapting these methods for driving applications remains challenging. To this end, we propose a systematic framework to enhance POMDP planning for AVs.

\textbf{Behavior Prediction}.
Recent advancements in learning-based prediction models have greatly improved accuracy in various benchmarks \cite{shi2023mtr++, nayakanti2023wayformer, zhou2023query}. However, when applying these models in online and closed-loop settings, prediction consistency becomes a major problem because these models are all designed to work with independent inputs and are trained offline. To address this problem, the motion forecasting with self-consistent constraints (MISC) method \cite{ye2023bootstrap} introduced self-consistent losses in training, but it still uses independent frames as inputs and is trained offline. The concept of streaming motion prediction \cite{pang2023streaming} has been proposed recently, where a neural differentiable filter is used to update the predicted states of agents at each timestep. Our method updates the latent belief state instead, which ensures performance and temporal consistency and allows for incorporating the AV's influence in the update. Moreover, many studies focus on ego-conditioned prediction (i.e., incorporating the future actions of the AV into the predictions of other agents) to make interaction-aware decisions \cite{huang2023dtpp, espinoza2022deep}. However, most of these models are trained in an open-loop manner. In contrast, our model is trained online and capable of modeling closed-loop interactions (\textcolor{red}{\textbf{C3}}).

\begin{figure*}
    \centering
    \includegraphics[width=0.82\linewidth]{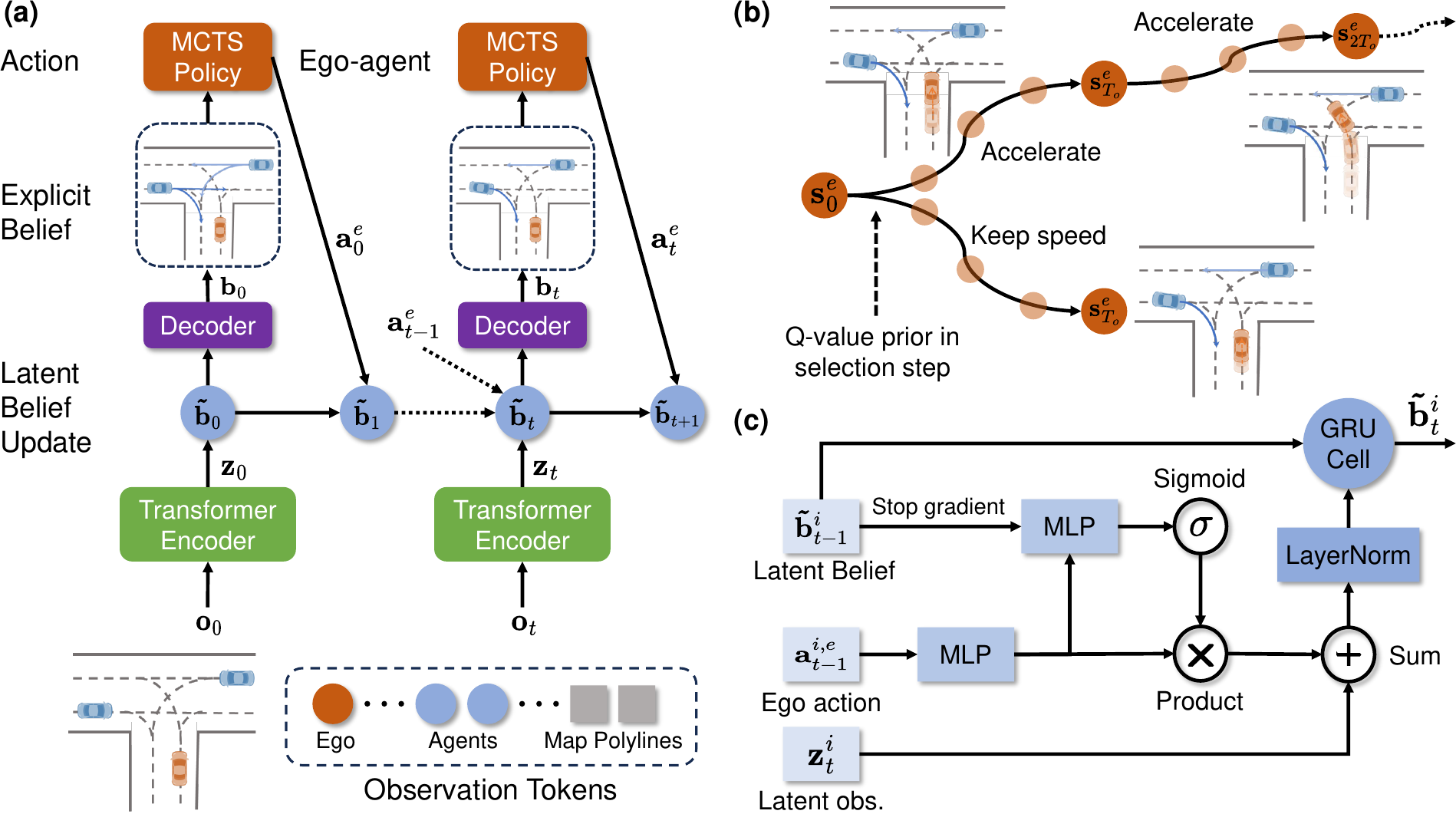}
    \caption{Illustration of the proposed POMDP decision-making framework. \textbf{(a)} Diagram of the POMDP planning process. At each time step, we use a Transformer encoder to map the observation into latent space, which is used to update the latent belief state. We use an MLP decoder to project the latent belief state into probabilistic future trajectories per agent. \textbf{(b)} Macro-action-based MCTS planner. The planner takes as input the estimated future states of other agents and searches over macro-actions. Additionally, we employ a learned Q-value network to guide the selection step at the root node. \textbf{(c)} Structure of the belief update network. The information of the previous latent belief state, current latent observation, and the ego agent's action are fused in the update process.}
    \label{fig:2}
    \vspace{-0.5cm}
\end{figure*}

\section{Method}
\subsection{Problem Statement}
We formulate the AV decision-making problem as a POMDP, which is defined by the tuple $(\mathcal{S}, \mathcal{A}, \mathcal{O}, \mathcal{Z}, \mathcal{T}, \mathcal{R})$. $\mathcal{S}$ is the state space and $\mathbf{s}_{t}^i$ denotes the state of traffic agent $i$ and time $t$. $\mathbf{a}_t^e \in \mathcal{A}$ is the action space for the ego agent. $[\mathbf{o}_{t}^i, \mathbf{o}_{t}^m] \in \mathcal{O}$ is the observation space, where $\mathbf{o}^{i}_t$ denotes the observed physical state of the agent (e.g., position, orientation, velocity, and size) and $\mathbf{o}^{m}_t$ is the shared road map. $\mathcal{Z}(\mathbf{o}_t | \mathbf{s}_{t})$ is the observation function, $\mathcal{T}(\mathbf{s}_{t+1} | \mathbf{a}_t, \mathbf{s}_{t})$ is the state transition function, and $\mathcal{R}(\mathbf{s}_t, \mathbf{a}_t)$ is the immediate reward function.

The ego (decision-making) agent cannot directly observe the internal state $\mathbf{s}^{i}$ of other agents, which can encompass attributes such as the agent’s navigational goals, behavioral traits, and intentions \cite{brown2020taxonomy}. Therefore, the ego agent maintains a \textit{belief} state $\mathbf{b}_t = \{\mathbf{b}_t^i \}_{i=1:N}$ as an estimate of the intentions of its surrounding $N$ agents. The belief state starts with an initial belief $\mathbf{b}_0^{i}$ and is updated $\mathbf{b}_t^{i}$ at each time step $t$ according to a belief update function $\rho$:
\begin{equation}
\mathbf{b}^i_{t} = \rho(\mathbf{b}^i_{t-1}, \mathbf{a}^e_{t-1}, \mathbf{o}_{t}^{i}),    
\end{equation}
where $\mathbf{a}^e$ is the action of the ego agent.

We assume that the ego agent makes decisions based on its current belief of the surrounding agents; thus, the entire history of observations is taken into consideration implicitly in the ego agent's actions. We aim to find the approximately optimal ego policy, denoted by $\pi(\mathbf{a}_t | \mathbf{b}_t)$, that maximizes the expected total discounted reward:
\begin{equation}
\pi^* = \arg \max_{\pi} \ \mathbb{E} \left[  \sum_{t=0}^{T} \gamma^t \mathcal{R} \left(\mathbf{s}_t, \pi(\mathbf{b}_t) \right) \right],
\end{equation}
where $T$ is the horizon and $\gamma \in (0, 1)$ is the discount factor. 

Specifically, we represent the state of an agent $\mathbf{s}^i$ as its intention or future plan. We maintain a \textit{latent belief} of the ego agent about the states of other agents, which can be decoded back to an \textit{explicit belief} state, such as probabilistic intentions or multi-modal future trajectories. To achieve this, we employ an encoder network that maps the observation to a latent space, and a GRU network to update the latent belief state, which is then decoded to possible trajectories per agent. The overall POMDP planning framework is illustrated in \cref{fig:2}.

\subsection{Neural Recurrent Belief Update}
\label{online model}
\textbf{Observation}. 
The observation space \cite{shi2023mtr++} typically includes the historical trajectories of agents (including the ego) $\mathbf{o}^a \in \mathbb{R}^{N_a \times T_h \times D_a}$ and map polylines  $\mathbf{o}^m \in \mathbb{R}^{N_m \times N_w \times D_m}$, where $N_a$ is the number of agents, $T_f$ is the fixed history steps, $N_m$ is the number of polylines, and $N_w$ is the number of waypoints in a polyline. Note that their positional attributes are normalized according to their respective origin coordinates. We encode the historical states of agents using GRU networks and multi-layer perceptron (MLP) networks to encode the map elements, and we apply max-pooling to reduce the waypoint axis. Finally, we concatenate these elements to obtain an encoding of the scene $\mathbf{h} \in \mathbb{R}^{(N_a+N_m) \times D}$, where $D$ is the hidden feature dimension.
 
Subsequently, we employ a Transformer network with query-centric multi-head attention \cite{shi2023mtr++, zhou2023query} to extract the relationships between these scene elements in a symmetric manner. Since each agent is encoded in their local coordinates rather than a global coordinate, it is invariant to the position change of the ego agent and thus facilitates belief update. Specifically, we apply pair-wise attention calculation for each query token from encoding $\mathbf{h}$, and the query, key, and value inputs of the attention module are derived as follows.
\begin{equation}
\begin{split}
Q^{i} &= \mathbf{h}^i, \  
K^{ij} = \mathbf{h}^j + \mathbf{e}^{ij}, \
V^{ij} = \mathbf{h}^j + \mathbf{e}^{ij}, \\
\mathbf{e}^{ij} &= PE(\Delta_{x}^{ij}, \Delta_{y}^{ij}, \Delta_{heading}^{ij}),
\end{split}
\end{equation}
where $i, j$ are the indexes of query tokens, and $PE$ stands for positional encoding, which is a feed-forward network.

We denote the encoding process as a function with parameters $f_{\theta}$ and the final encoding as $\mathbf{z} \in \mathbb{R}^{(N_a+N_m) \times D} $, and we can retrieve $N$ agents of interest from the scene encoding.

\textbf{Belief Update}. 
We denote the belief update network as $f_\rho$, which takes as input the last latent state $\mathbf{\tilde{b}}_{t-1}^i \in \mathbb{R}^{D}$, the current encoded observation $\mathbf{z}_{t}^i$, and the ego agent's planned actions (short-term trajectory) relative to each agent $\mathbf{a}_{t-1}^{i,e} \in \mathbb{R}^{T_o \times 3}$. We define the initial latent belief as $\mathbf{\tilde{b}}_{0}^i \triangleq \mathbf{z}_{0}^{i}$, and the belief update process can be formulated as follows.
\begin{equation}
\begin{split}
\mathbf{z}_{t}^i &= f_{\theta} (\mathbf{o}_t^i, \mathbf{o}_t^{\neg i}, \mathbf{o}_{t}^m), \\
\mathbf{\tilde{b}}_{t}^{i} &= f_{\rho} (\mathbf{z}_t^i, \mathbf{\tilde{b}}_{t-1}^i, \mathbf{a}_{t-1}^{i,e}), \\
\mathbf{b}^i_t &= f_{\phi} (\mathbf{\tilde{b}}_{t}^{i}),
\end{split}   
\end{equation}
where $\mathbf{b}^i_t$ represents the explicit belief state, and $f_\phi$ is the trajectory decoder network that decodes multi-modal predictions from the latent belief state.

The structure of the belief update network is illustrated in \cref{fig:2}(c). We first account for the influence of the ego agent's action on the state of the tracked agent. For this, we use two MLPs - one is utilized to encode the ego agent's relative trajectory, and the other is used to predict a gating value. We use this gating value to control the flow of information to the belief update because some agents may not be influenced by the ego agent. Afterward, the ego agent's influence is combined with the encoded observation to update the belief state through the GRU cell. Note that some agents may be occluded for a few timesteps, and we take the last available belief state of those agents in the update.

\textbf{Trajectory Prediction}. 
Given the latent belief state about an agent $\mathbf{\tilde{b}}_{t}^{i}$, we use an MLP $f_\phi$ to decode from the latent belief to explicit belief, which is represented by a Gaussian mixture model (GMM) of future trajectories $\mathbf{b}^i \in \mathbb{R}^{M \times T_f \times D_f}$. This GMM comprises $M$ modalities and $T_f$ future timesteps, and features at each step include $x$ and $y$ coordinates, variances $\sigma_x$ and $\sigma_y$, and probability $p$. It is important to note that the decoder $f_\phi$ can also be used for the encoded observation $\mathbf{z}^i$, as in most existing offline motion prediction models.

\subsection{Macro-action-based MCTS Planner}
\label{mcts}
We use the belief state of the ego agent over other agents $\mathbf{b}_{t} = \{ \mathbf{b}_{t}^{i} \}_{i=1:N}$ to approximate the future updates within the planning horizon (multi-modal long-term trajectories). This effectively simplifies the POMDP problem into a belief-state MDP \cite{sunberg2022improving}, which can be efficiently solved using the MCTS algorithm. We employ receding horizon planning and construct a new policy for the current belief each time a new observation is received. Furthermore, to optimize the quality and efficiency of the MCTS planner, we employ macro-actions.

\textbf{Macro-actions}.
A macro-action is a sequence of instantaneous actions over a time duration \cite{amato2019modeling, de2016monte}, and they can be learned from data and differ in length \cite{zhou2023towards}. For simplicity, we employ fixed-length heuristic macro-actions (high-level discrete intentions, such as overtaking and turning). We assume that the routing module provides a reference path, such that we can generate macro-actions based on this reference path. At each node, macro-actions are generated using the available actions for acceleration ($a_s$) and lateral speed ($v_l$), and the resulting states are calculated in the Frenet frame: 
\begin{equation}
\footnotesize
\begin{split}
&\ddot s(\tau) = a_s, \ \dot s(\tau) = \dot s(0) + a_s \tau, \ s(\tau) = s(0) + \dot s(0)\tau + \frac{1}{2} a_s \tau^2,  \\
&\dot l(\tau)  = v_l, \ l(\tau) = l(0) + v_l \tau, \ \tau \in [1, 2, \cdots, T_o] \Delta \tau, 
\end{split}
\end{equation}
where $s$ and $l$ represent the longitudinal and lateral directions along the reference path, $s(0)$, $l(0)$, $\dot s(0)$ are the initial state of the node, $\Delta \tau$ is the time interval, and $T_o$ is the macro-action length. Then, we can transform the states in the Frenet frame back into the Cartesian frame $\left( x(\tau), y(\tau), v(\tau), a(\tau) \right)$ \cite{zhang2020optimal}.

\textbf{Objective Function}.
We calculate the cost of macro-actions considering collision risk, ride comfort, deviation from the route, and desired speed. The objective function is:
\begin{equation}
\max \sum_{n=0}^{T_d-1} \gamma^{n T_o \Delta \tau} \sum_{i} - w_i c_i(n),     
\end{equation}
where $T_d = \frac{T_f}{T_o}$ is the search depth, $w_i$ is the weight, and $c_i$ is the cost term at time segment $n$. The ride comfort factor includes acceleration, jerk, and lateral acceleration of the trajectory, and the desired speed cost is the difference between the current speed and speed limit. Each cost term is averaged across all timesteps in the node.

Importantly, the collision risk of an option is calculated as:
\begin{equation}
\small
\label{pth}
c_{r}(n) = \sum_{i}^{N} \sum_{j}^{M} \mathds{1}(p_j \geq p_{th}) p_j \mathds{1}_{overlap} \left( \mathbf{s}^e(n), \mathbf{b}^{i}(j, n) \right),
\end{equation}
where $\mathbf{s}^e$ is the state of the ego agent determined by the generated macro-actions, $\mathbf{b}^{i}(j)$ is the $j$-th predicted trajectory of agent $i$, $p_j$ is the predicted likelihood of that trajectory. $ \mathds{1}_{overlap}$ is a binary indicator to check whether the two trajectories collide, and $p_{th}$ is a threshold to filter out low probability futures.

\renewcommand{\Comment}[2][gray]{\textcolor{#1}{\hfill// #2}}

\begin{algorithm}
    \caption{MCTS with online belief prediction model}
    \begin{algorithmic}[1]
    \Require MCTS planner $TP$, observation encoder $f_\theta$, belief update network $f_\rho$, trajectory decoder $f_\phi$, Q-value network $Q_\epsilon$, $N$ number of predicted agents.
    
    \State Initialize timestep $t \gets 0$ and belief state tracker $B \gets \O$
    \While{task is active}
        \State Observe environment $\mathbf{o}_t = \left[ \{ \mathbf{o}_{t}^i\}_{i=1:N_a}, \mathbf{o}^m_t \right]$ 
        \State Encode observation and retrieve $N$ agent tokens 
        \Statex \hspace{0.5cm} of interest $\{\mathbf{z}_{t}^i\}_{i=1:N} = f_{\theta} (\mathbf{o}_t)$ 
        \For{$i \gets 1$ to $N$}
            \If{$\mathbf{\tilde b}_{t-1}^i \in B$}
                \State $\mathbf{\tilde b}_{t}^{i} = f_{\rho} (\mathbf{z}_t^i, \mathbf{\tilde b}_{t-1}^i, \mathbf{a}_{t-1}^{i,e}) $\Comment{Update belief state}
            \Else
                \State $\mathbf{\tilde b}_{t}^i = \mathbf{z}_t^i$ \Comment{Initialize belief state}
            \EndIf
            \State Update state tracker $B \gets B \cup \{\mathbf{\tilde b}_t^i\}$
            \State Decode probabilistic trajectories $\mathbf{b}_t^i = f_\phi (\mathbf{\tilde b}_t^i)$
        \EndFor
        \State Calculate prior $Pr(\mathbf{z}_t^e, \mathbf{a}^e)$ for root node using \cref{prior}
        \State Compute action sequence ${\mathbf{a}}^* \gets TP(\{\mathbf{b}_t^i\}_{i=1:N}, Pr)$ 
        \State Execute first action ${\mathbf{a}_{t}^{e}} \gets \mathbf{a}_{0}^*$ 
        \State Increment timestep $t \gets t+1$
    \EndWhile
    \end{algorithmic}
    \label{algo1}
\end{algorithm}

\textbf{MCTS Algorithm}. 
MCTS consists of several key steps: selection, expansion, evaluation, and backup, and this process is repeated multiple times until a termination condition is reached. We focus on the selection step in the following, and more details about MCTS can be found in \cite{chekroun2023mbappe}. In the selection step, the nodes are selected according to:
\begin{equation}
a = \arg \max_{a \in A} \tilde Q(s, a) + Pr(a) C_p \sqrt{\frac{2 \log N_p}{n_a + 1}} , 
\end{equation}
where $\tilde Q(s, a)$ is the expected return, $n_a$ is the number of times the node has been visited, $N_p$ is the total number of times the parent node has been visited, and $C_p$ is a temperature parameter used to balance exploration and exploitation.

\textbf{DQN Guidance}.
DQN is used in MCTS as an action prior to bias exploring high-rewarding areas in the search tree. The prior $Pr(a)$ is given by the learned Q-value network $Q_\epsilon$, and we apply this prior to the root node, while setting the prior to a uniform distribution in other nodes. The sampling prior for the root node is calculated as:
\begin{equation}
\label{prior}
Pr(a_i) = \frac{\exp Q_{\epsilon}(\mathbf{z}^e_t, a_i)}{\sum_j \exp Q_{\epsilon}(\mathbf{z}^e_t, a_j)},
\end{equation}
where $\mathbf{z}^e_t$ is the encoded observation state for the ego agent at the current time, which is retrieved from the observation encoder and contains necessary information including its past states, other agents, and road maps. 

The proposed POMDP deep predictive planning algorithm can be summarized in \cref{algo1}.

\subsection{Learning Framework}
\label{learning}
\textbf{Offline Learning}.
In this stage, we train the observation encoder $f_\theta$ with the multi-modal trajectory decoder $f_\phi$ using an offline driving dataset. We employ the following loss on the selected positive GMM component (the mode with the smallest L2 distance to ground truth) for a specific agent at a future time step. 
\begin{equation}
\footnotesize
\label{gmm}
\mathcal{L}_{GMM} =  \log \hat \sigma_x + \log \hat \sigma_y + \frac{1}{2} \left( \left(\frac{s_x - \hat s_x}{\hat \sigma_x}\right)^2 + \left(\frac{s_y - \hat s_y}{\hat \sigma_y}\right)^2 \right) \! - \log \hat p,
\end{equation}
where ($\hat s_x, \hat s_y, \hat \sigma_x, \hat \sigma_y, \hat p$) is the selected Gaussian component; $s_x, s_y$ are the ground-truth future states of the predicted agent.

\begin{algorithm}
    \caption{Online learning process}
    \begin{algorithmic}[1]
    \Require pre-trained observation encoder $f_\theta$, belief update network $f_\rho$, trajectory decoder $f_\phi$, twin Q-value network $Q_\epsilon^{1, 2}$, training steps $N_t$, exploration rate $\lambda$ 
    \State Initialize the replay buffer $R$
    \For{$i \gets 1$ to $N_t$}
    \State Initialize episodic observation buffer $O$ and trajectory 
    \Statex \hspace{\algorithmicindent} prediction buffer $P$
    \While{episode not terminated}
        \State Collect and store agent observations in $O$
        \State Encode observations to latent space $z$ using $f_\theta$
        \State Predict trajectories using $f_\rho$ and $f_\phi$ and store in $P$
        \State Select action $a$: with probability $\lambda$, sample random 
        \Statex \hspace{1.2cm} action; otherwise use MCTS planner
        \State Execute action $a$, observe reward $r$, and obtain
        \Statex \hspace{1cm} next latent state $z'$
        \State Store transition $(z, a, r, z')$ in replay buffer $R$
        \State Update $Q_\epsilon^{1, 2}$ by sampling from $R$
    \EndWhile
    \State Update $f_\rho$ and $f_\phi$ using prediction results from $P$ and 
    \Statex \hspace{\algorithmicindent} ground truth data from $O$ in hindsight
    \EndFor
    \State \Return Trained networks $f_\rho$, $f_\phi$, $Q_\epsilon^{1}$
    \end{algorithmic}
    \label{algo2}
\end{algorithm}

\textbf{Online Learning}.
In this stage, we fix the weights of the observation encoder and learn the belief update model $f_\rho$ and fine-tune the trajectory decoder $f_\phi$ in a driving simulator. At the end of an episode, we obtain a prediction buffer and an observation buffer that contains the positions of all agents at each timestep. We utilize the GMM loss in \cref{gmm} to minimize the prediction loss at each valid time step in hindsight. In addition, we leverage the clipped double Q-learning \cite{fujimoto2018addressing} to learn a Q-value network $Q_\epsilon$ to guide the tree search planner. The reward function from the environment for Q-learning is formulated as:
\begin{equation}
r = r_{col} + r_{prog} - 0.01 r_{expert},    
\end{equation}
where $r_{col}=-10$ is the penalty imposed if the ego vehicle collides with any other agents, $r_{prog}$ is the distance traveled by the ego vehicle, and $r_{expert}$ is the state difference between the agent and logged human expert, which could encode other driving features that are hard to manually design. Since the true state of the environment can only be obtained at the current step, and future states are approximated by the model, the Q-value network is used to guide the selection at the root node.

We jointly learn the belief update and Q-network (with the MCTS planner) to improve the overall decision-making performance. The procedure for online belief update model learning and Q-learning is summarized in \cref{algo2}.

\section{Experiments}
\subsection{Dataset and Simulator}
\begin{figure}[htp]
    \centering
    \includegraphics[width=\linewidth]{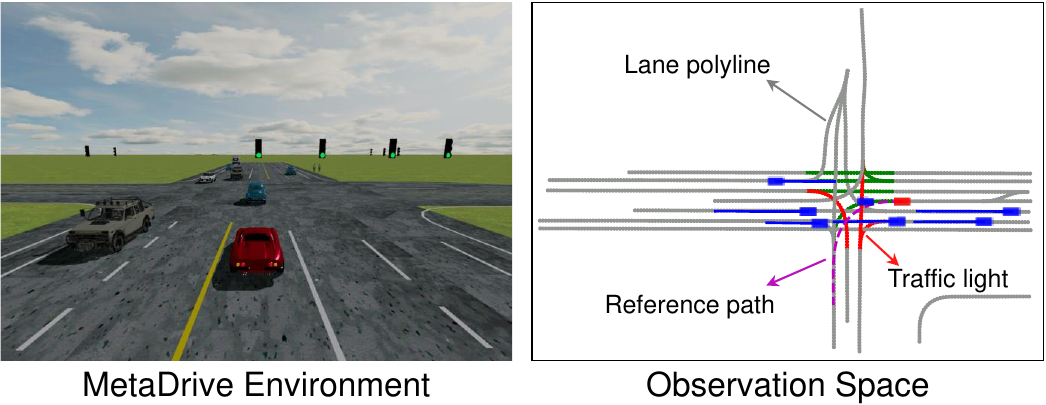}
    \caption{Example of replayed WOMD scenario in the MetaDrive simulator and the processed observation space including vectorized map polylines and historical trajectories of surrounding agents. }
    \label{fig:3}
    \vspace{-0.6cm}
\end{figure}

We employ the Waymo Open Motion Dataset (WOMD) \cite{ettinger2021large} to train the models and replay the traffic data for surrounding agents in the MetaDrive \cite{li2022metadrive} simulator to conduct online training and testing. Additionally, we set the agent behavior to be reactive in the simulator, allowing other agents to perform basic reactive actions, such as avoiding collisions from the rear end of the ego vehicle. \cref{fig:3} shows a replayed scenario from WOMD in the MetaDrive simulator and the obtained observation space for the ego agent.

\subsection{Implementation Details}
\textbf{Data}. 
In the offline learning stage, we select 20,000 scenarios from the WOMD, each with a duration of 20 seconds. For online learning, we randomly choose 2,000 scenarios from the training set for simulation-based replay. In the testing phase, we select an additional 200 scenarios from the WOMD. Note that we exclude static scenarios where the ego vehicle travels less than 20 meters, such as waiting at red lights.

\textbf{Neural Networks}.
 We select the closest $N_a=16$ agents to the ego vehicle and their observed historical horizon is $T_h=20$ timesteps. Additionally, we incorporate $N_m=50$ map polylines in encoding, each comprising $N_w=20$ waypoints. The Transformer encoder has $3$ attention layers and a hidden dimension of $D=256$. The trajectory decoder outputs $M=6$ possible trajectories for an agent, $8$ seconds into the future $T_f = 80$. In the belief update network, the ego agent's action is represented by a short-term trajectory ($T_o=20$), which is processed relative to each tracked agent and encoded. The Q-value network is an MLP that takes as input the hidden state of the ego agent and outputs the values of actions.

\textbf{MCTS Planner}.
The planning horizon is set to $T_f=80$ timesteps, with a time interval of $\Delta t = 0.1 \ s$, and the option length is set to $T_o = 20$. This significantly reduces the search depth to $T_d = 4$. We track the belief states of $N=8$ agents and set the probability threshold for calculating collision risk to $p_{th} = 0.15$. The options at each node are derived from combinations of lateral velocity $v_l = [-1, 0, 1] \ m/s$ and longitudinal acceleration $a = [-4, -2, 0, 1, 3] \ m/s^2$. Lateral velocity is only considered if the acceleration is $-2, 0, 1$, leading to a total of $11$ options. We also ensure that the velocity remains non-negative. The planner executes a total of $100$ search iterations and applies a discount factor $\gamma=0.8$ and temperature parameter $C_p=100$ after careful tuning.

\textbf{Offline learning}.
We utilize the AdamW optimizer for model training, conducting $40$ training epochs. The learning rate begins at $2e-4$ and is halved every $5$ epochs. The model is trained on an NVIDIA A100 GPU with a batch size of 256.

\textbf{Online Learning}.
We train the models in the simulator with $N_t = 300,000$ steps. We employ Adam optimizer to train the networks and the learning rate starts with $3e-4$ and reduces by $30\%$ every $50,000$ steps. The exploration rate $\lambda$ starts with $0.8$ and decays to $0.05$ in the middle of training. The discount factor for Q-learning is $\gamma=0.99$, the replay buffer size is $100,000$, and the batch size is $128$.

\textbf{Testing}.
The runtime of the prediction model and MCTS planner is tested on a laptop equipped with an AMD 7945HX CPU and NVIDIA RTX 4060 GPU.

\subsection{Main Results}
\textbf{Baselines}.
In testing, we compare our method against several strong baseline approaches: RL (DQN) method \cite{fujimoto2018addressing}, intelligent driver model (IDM) \cite{zhang2022bayesian}, and differentiable integrated prediction and planning (DIPP) \cite{huang2023differentiable}. We also integrate well-established trajectory prediction models from the WOMD benchmark, such as MTR \cite{shi2023mtr++} and GameFormer \cite{huang2023gameformer}, into the MCTS planner to test their performance on planning. To ensure a fair comparison, we scale down these models to match the size of ours and train them using the same dataset.

Moreover, we introduce several ablated versions of our approach. The \textbf{offline model} denotes the prediction model that operates without online belief updates (i.e., decoding trajectories directly from encoded observations), and is only trained on the offline dataset. The proposed \textbf{online update model} has two variants: one incorporates the ego agent's intentions in the online belief update, and the other does not. Furthermore, we evaluate the effectiveness of employing imitation learning (IL) (i.e., predicted future trajectory for the ego vehicle) \cite{chekroun2023mbappe, li2022efficient} and DQN as prior in the MCTS planner. For training comparisons, we introduce two baselines: a DQN agent and an MCTS planner that includes an online belief update model but omits Q-learning.

\textbf{Metrics}. 
\newcommand{\norm}[1]{\left\lVert#1\right\rVert}
The task-related metrics include success rate, average episode reward, task time (defined as the average number of steps per episode), and average log divergence (L2 distance) measuring the deviation of the ego vehicle from logged behavior. Regarding prediction performance, we employ three key metrics: minimum average displacement error (minADE), consistency (variation in predictions across consecutive frames), and score accuracy (whether the highest probability is correctly assigned to the closest modality to the ground truth). Prediction consistency is defined as:
\[
\frac{1}{N} \sum_{i=1}^N \frac{1}{M} \sum_{m=1}^{M} \frac{1}{T_f-1} \sum_{\tau=1}^{T_f-1} \norm{\mathbf{b}^i_{t}(m, \tau) - \mathbf{b}^i_{t-1}(m, \tau+1) }^2,
\]
where $\mathbf{b}_{t-1}$ is the prediction result from the last time frame, and this metric is averaged across all timesteps in the episode.

\begin{figure}[htp]
    \centering
    \includegraphics[width=0.95\linewidth]{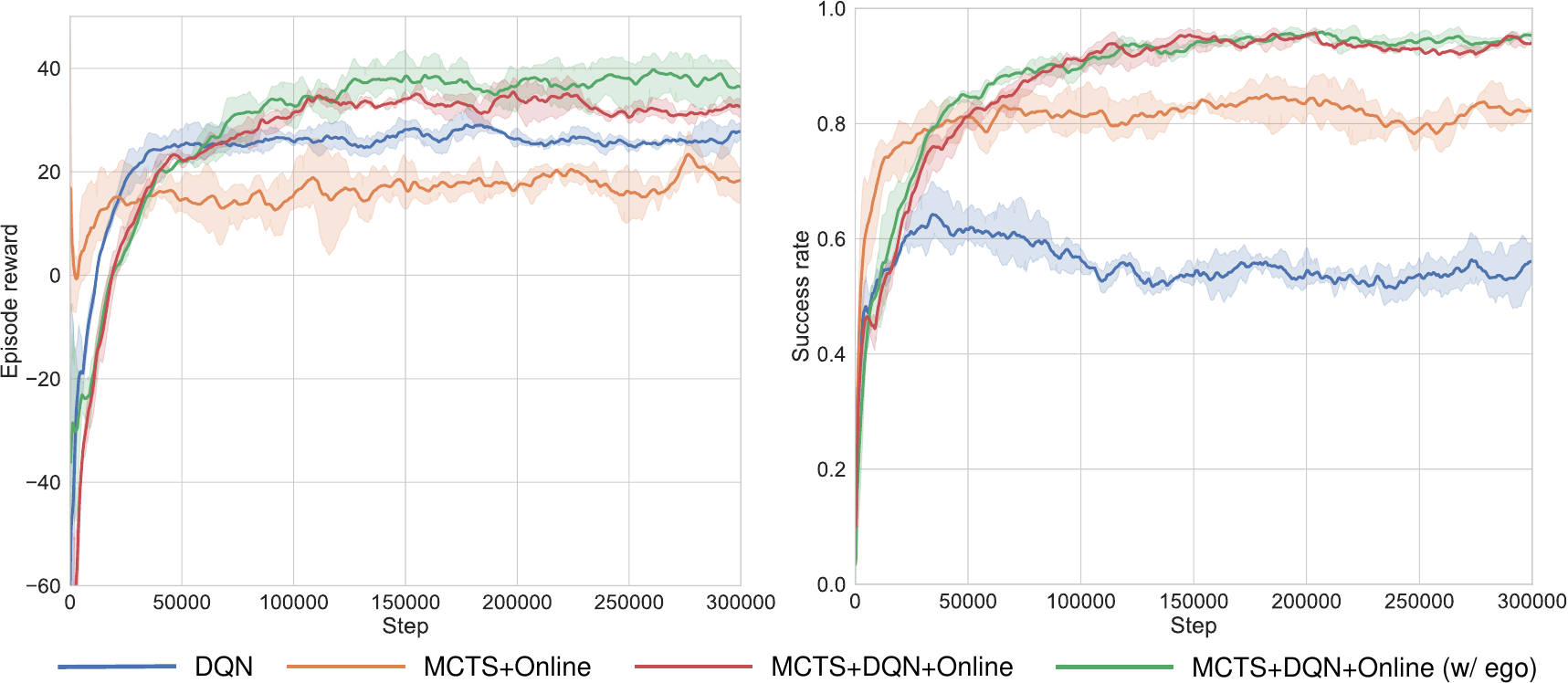}
    \caption{Training results of our proposed method against baseline methods. \textbf{Left}: average episodic reward; \textbf{right}: average success rate.}
    \label{fig:4}
    \vspace{-0.3cm}
\end{figure}

\textbf{Training Results}. 
We train different methods using three random seeds, and the training results are depicted in \cref{fig:4}. The DQN agent shows the worst performance in terms of success rate, and learning the online prediction model without Q-learning significantly influences the MCTS planner's effectiveness, mainly due to the deviation from expert trajectories. While the MCTS planner with an online prediction model and deep Q-learning shows favorable effectiveness, its performance can be further enhanced with the online model that integrates the ego agent's intentions into the belief update process. This approach achieves the highest reward, highlighting the efficacy of closed-loop interaction modeling.

\begin{table*}[htp]
\centering
\caption{Performance comparison of different decision-making approaches in the testing scenarios}
\label{tab:1}
 \resizebox{0.92\textwidth}{!}{
\begin{tabular}{@{}l|cccc|ccc@{}}
\toprule
Method                              & Success rate                  & Task time                 & Log divergence            & Episode reward            & Pred. minADE  ($\downarrow$) & Consistency ($\downarrow$) & Score accuracy ($\uparrow$) \\ \midrule
IDM Agent                           & 0.91                          & 132.18                    &  11.80                    & 38.59                     &  --               & --             & -- \\
DIPP \cite{huang2023differentiable} & 0.94                          & 125.65                    &  10.35                    & 43.88                     & 0.902             &  1.631         & 0.260 \\
DQN Agent                           & 0.53$\pm$0.04                 &\textbf{62.32}$\pm$10.74   &  11.87$\pm$0.58           & 30.08$\pm$1.35            &   --              &  --            & --    \\ 
MCTS + DQN + MTR \cite{shi2023mtr++}& 0.94$\pm$0.01                 & 120.15$\pm$1.15           &  10.48$\pm$0.16           & 45.55$\pm$0.74            & 0.841             &  1.405         & 0.281 \\ 
MCTS + DQN + GameFormer \cite{huang2023gameformer} & 0.94$\pm$0.01  & 119.85$\pm$1.05           &  10.47$\pm$0.11           & 45.68$\pm$0.34            & 0.833             &  1.682         & 0.265 \\ \midrule
MCTS + Offline model                & 0.78$\pm$0.02                 & 174.20$\pm$2.12           &  10.43$\pm$0.10           & 17.86$\pm$0.80            & 1.025             &  1.652         & 0.261\\
MCTS + IL + Offline model           & 0.72$\pm$0.04                 & 175.83$\pm$1.37           &  10.89$\pm$0.16           & 16.84$\pm$1.17            & 1.019             &  1.674         & 0.264\\
MCTS + Online update                & 0.82$\pm$0.01                 & 168.29$\pm$1.08           &  10.59$\pm$0.24           & 21.28$\pm$0.75            & 1.047             & 0.574          & 0.314 \\
MCTS + IL + Online update           & 0.87$\pm$0.02                 & 141.06$\pm$1.65           &  10.57$\pm$0.16           & 31.86$\pm$0.84            & 0.984             & 0.573          & 0.315 \\ \midrule
MCTS + DQN + Offline model          & 0.94$\pm$0.01                 & 119.58$\pm$0.41           & 10.43$\pm$0.10            & 45.31$\pm$0.52            & 0.847             & 1.712          & 0.266 \\
MCTS + DQN + Online update          &\textbf{0.98}$\pm$0.02         & 122.24$\pm$1.23           & 10.35$\pm$0.07            & 46.21$\pm$0.18            & \textbf{0.829}    & 0.570          & 0.320  \\
MCTS + DQN + Online update (w/ ego) &\textbf{0.98}$\pm$0.02         & 119.50$\pm$0.94           & \textbf{10.27}$\pm$0.12   & \textbf{46.62}$\pm$0.26   & 0.878             & \textbf{0.562} & \textbf{0.323}  \\ \bottomrule
\end{tabular}
}
\vspace{-0.3cm}
\end{table*}

\begin{figure*}
    \centering
    \includegraphics[width=0.95\linewidth]{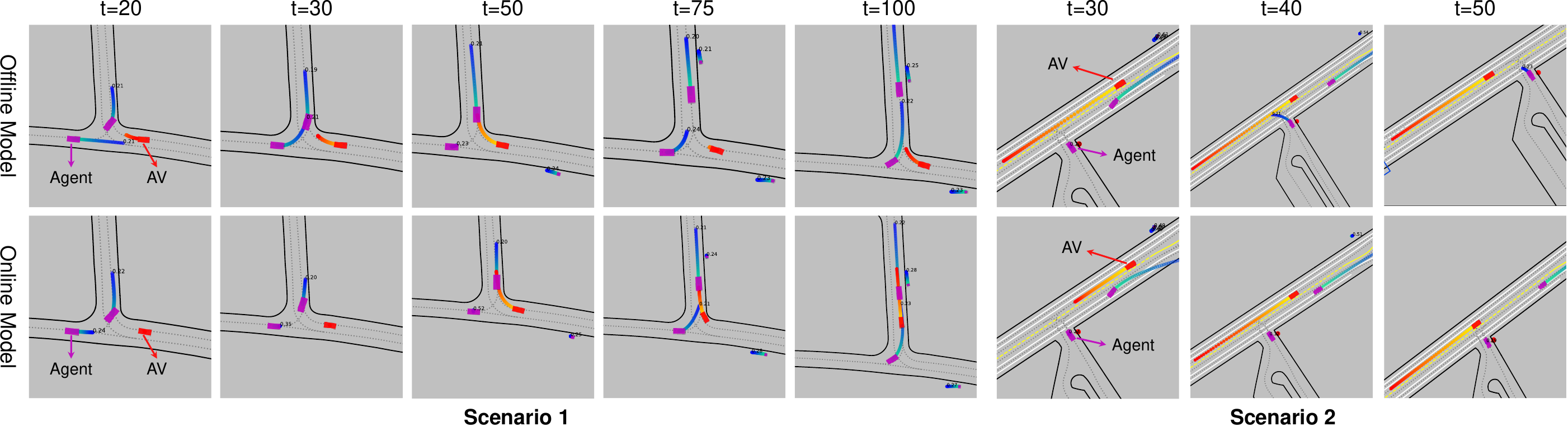}
    \caption{Qualitative comparisons of different behavior prediction models and their influences on the decision-making task. For clarity, only the most likely modality, along with its predicted probability, is displayed, and the length of trajectories is 5 seconds. The offline prediction model shows significant fluctuations in trajectories and probabilities, which cause the AV to fail to move ahead of the interacting agent in Scenario 1. In contrast, our online update model shows consistent predictions over time, with the accuracy gradually increasing when receiving new observations. The enhancements in behavior prediction provided by our online update model enable the MCTS planner to make more human-like decisions.}
    \label{fig:5}
    \vspace{-0.5cm}
\end{figure*}

\textbf{Testing Results.} 
We evaluate our proposed method against other baselines in the testing scenarios, and the results are summarized in \cref{tab:1}. The key conclusions and findings are as follows. \textbf{(1)} The DQN agent shows the weakest performance, while the rule-based IDM agent and the optimization-based DIPP method are more effective. Using established prediction models with our DQN-guided MCTS planner shows comparable performance to our offline-trained model but is less effective in planning than the online prediction model. \textbf{(2)} The basic MCTS method without DQN guidance performs significantly worse, and DQN guidance is more effective than IL guidance. \textbf{(3)} The prediction consistency of offline models is substantially worse than that of online belief update models. While offering similar prediction accuracy, the online belief update model significantly enhances prediction consistency and the accuracy of intention probability estimations, leading to improved decision-making performance. \textbf{(4)} Incorporating the ego vehicle's intention into the belief update process in the online prediction model can further improve the decision-making performance in terms of the episode reward. Some qualitative results are presented in \cref{fig:5}, which reveal that our proposed online update model provides more consistent and accurate trajectory predictions of interacting agents compared to the offline prediction model. The online prediction model can continuously improve its prediction accuracy (in terms of trajectory ADE and probability) by receiving new observations over time. The \href{https://drive.google.com/file/d/10gPa_EaCjcQpwL7bBoWfUX2HO6-jeWDu/view?usp=sharing}{closed-loop testing results} showcase the online prediction model's much better prediction consistency.

\subsection{Ablation Study}
\textbf{Influence of macro-action length}.
We investigate the impact of varying macro-action lengths on the performance of the MCTS planner, as detailed in \cref{tab:2}. We find that shorter action lengths (1, 5, 10) lead to increased computation time and lower success rate and episode reward. This is because, within a limited number of search iterations, a deeper search depth makes it increasingly difficult to find better actions. Conversely, extending the macro-action length to 40, which essentially simplifies the process into two-stage planning, yields a success rate comparable to a macro-action length of 20, but the episode reward at this length is reduced. Consequently, a macro-action length of 20 appears to provide a good balance between performance and computational efficiency.

\begin{table}[htp]
\vspace{-0.2cm}
\centering
\caption{Influence of option length on planning performance}
\label{tab:2}
\resizebox{0.9\linewidth}{!}{
\begin{tabular}{@{}l|cccc@{}}
\toprule
Length  & Success rate  & Episode reward & Log divergence & Runtime ($ms$)  \\ \midrule
1       & 0.59          & 21.47          & 14.34          & 834.8 \\
5       & 0.62          & 25.90          & 13.58          & 376.5 \\
10      & 0.88          & 37.12          & 12.08          & 196.4 \\
20      & \textbf{0.98} & \textbf{46.62} & \textbf{10.27} & 126.5 \\
40      & \textbf{0.98} &  42.25         & 10.53          & \textbf{92.4} \\ \bottomrule
\end{tabular}
}
\vspace{-0.2cm}
\end{table}

\textbf{Influence of probability threshold}.
We examine the effects of the probability threshold in the collision risk cost (\cref{pth}) on planning performance. Setting $p_{th}=0$ includes all predicted futures in planning, and we also set up an alternative approach that only considers the most likely predicted future. The results in \cref{tab:3} reveal that incorporating all uncertainties yields a high success rate but a lower episode reward compared to $p_{th}=0.15$. This indicates that while accounting for all uncertainties ensures task completion, it often results in deviations from the human trajectory and thus decreased rewards. Utilizing the most likely future leads to suboptimal performance in terms of both success and reward, highlighting the limitations of optimistic planning strategies.

\begin{table}[htp]
\centering
\vspace{-0.2cm}
\caption{Influence of probability threshold on planning performance}
\label{tab:3}
\begin{tabular}{@{}l|ccc@{}}
\toprule
Threshold   & Success rate  & Episode reward & Log divergence \\ \midrule
0           & \textbf{0.98} & 44.85          & 10.49          \\
0.15        & \textbf{0.98} & \textbf{46.62} & \textbf{10.27}   \\
$p_{\max}$    & 0.94          & 44.28          & 10.39          \\ \bottomrule
\end{tabular}
\vspace{-0.4cm}
\end{table}

\section{Conclusions}
We develop an online behavior prediction model for POMDP planning in autonomous driving. Specifically, we propose a recurrent neural belief update model and a macro-action-based MCTS planner guided by deep Q-learning. We introduce an online learning framework, which combines belief update learning and deep Q-learning to guide tree search. We validate our framework in simulated environments based on real-world driving scenarios. The experimental results indicate that our proposed online belief update model can significantly improve temporal consistency and accuracy. Furthermore, DQN guidance considerably boosts the efficacy of the MCTS planner, resulting in improved decision-making performance. Future research could extend this framework to broader POMDP settings by learning to update both actual physical states and future intentions. Moreover, the macro-action generation process can be improved by learning from expert data and by making the macro-actions more flexible in terms of their length, manipulation, and termination criteria.

\textbf{Acknowledgement}.
This work was supported by Berkeley DeepDrive. We would like to extend our gratitude to Yixiao Wang from UC Berkeley for his insights and support, as well as Rouhollah Jafari and Stefano Bonasera from General Motors for their constructive feedback.
\vspace{-0.2cm}

\bibliographystyle{IEEEtran}
\bibliography{IEEEexample}
\end{document}